\documentclass[letterpaper, 10 pt, conference]{ieeeconf}  
\IEEEoverridecommandlockouts                              
\usepackage{packages}

\title{\LARGE\bf 

Model Predictive Control with Visibility Graphs for Humanoid\\ Path Planning and Tracking Against Adversarial Opponents

}
\author{Ruochen Hou$^{1}$, Gabriel I.~Fernandez$^{1}$, Mingzhang Zhu$^{1}$, and Dennis W.~Hong$^{1}$
\thanks{
This paper has been accepted to IEEE International Conference on Robotics and Automation (ICRA) 2025. $^{1}$Robotics and Mechanisms Laboratory (RoMeLa), Department of Mechanical and Aerospace Engineering, University of California, Los Angeles, CA 90095, USA.
        {\tt\small \{houruochen, gabriel808, normanzmz, dennishong\}@ucla.edu}}
}

\begin{document}
\maketitle
\thispagestyle{empty}
\pagestyle{empty}

\begin{abstract}
In this paper we detail the methods used for obstacle avoidance, path planning, and trajectory tracking that helped us win the adult-sized, autonomous humanoid soccer league in RoboCup 2024. Our team was undefeated for all seated matches and scored 45 goals over 6 games, winning the championship game 6 to 1. During the competition, a major challenge for collision avoidance was the measurement noise coming from bipedal locomotion and a limited field of view (FOV). Furthermore, obstacles would sporadically jump in and out of our planned trajectory. At times our estimator would place our robot inside a hard constraint. Any planner in this competition must also be be computationally efficient enough to re-plan and react in real time. This motivated our approach to trajectory generation and tracking. In many scenarios long-term and short-term planning is needed. To efficiently find a long-term general path that avoids all obstacles we developed DAVG (Dynamic Augmented Visibility Graphs). DAVG focuses on essential path planning by setting certain regions to be active based on obstacles and the desired goal pose. By augmenting the states in the graph, turning angles are considered, which is crucial for a large soccer playing robot as turning may be more costly. A trajectory is formed by linearly interpolating between discrete points generated by DAVG. A modified version of model predictive control (MPC) is used to then track this trajectory called cf-MPC (Collision-Free MPC). This ensures short-term planning. Without having to switch formulations cf-MPC takes into account the robot dynamics and collision free constraints. Without a hard switch the control input can smoothly transition in cases where the noise places our robot inside a constraint boundary. The nonlinear formulation runs at approximately 120 Hz, while the quadratic version achieves around 400 Hz.

\end{abstract}


\section{Introduction}
\label{sec:intro}
\begin{figure}[t!]
    \centering
    \includegraphics[width=0.9\linewidth]{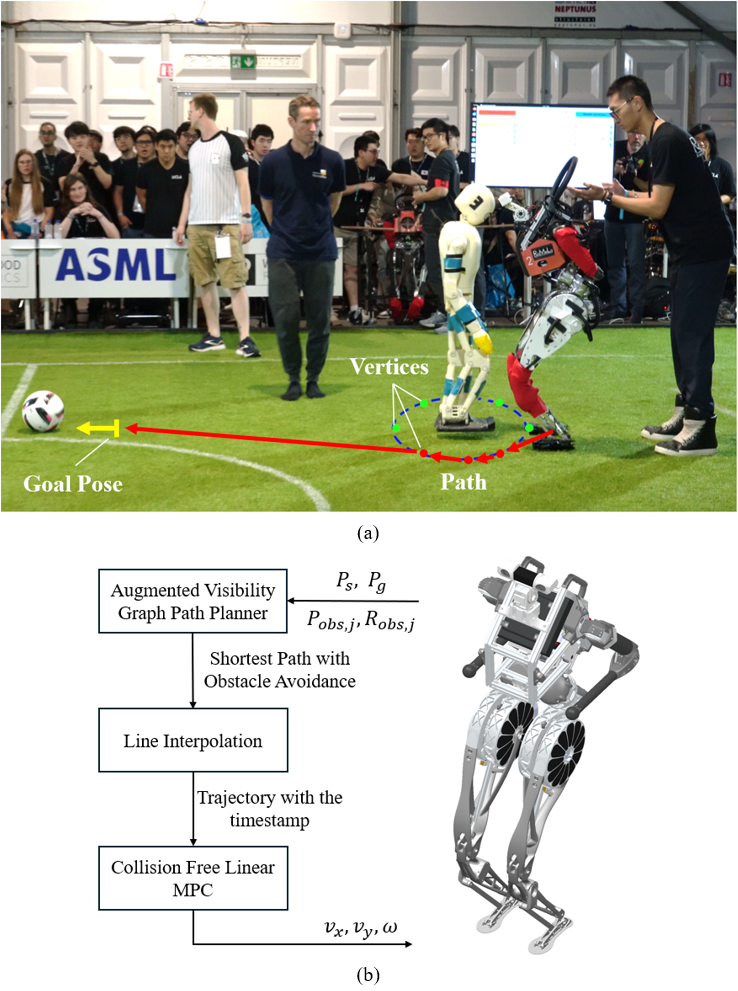}
    \caption{\textit{Top} figure shows a situation where our robot, ARTEMIS, in red is trying to avoid the opponent in blue to get to the ball during the championship match in RoboCup 2024. There are many situations like this or even worse when trying to reach the ball. \textit{Bottom} figure gives an overview of our entire framework for trajectory generation and tracking. $P_s$ and $P_g$ represent the starting pose and goal pose, respectively. $P_{obs,j}$ and $R_{obs,j}$ represent the position and radius of obstacles, respectively.}
    \label{fig:complete_process}
\end{figure}

Finding a collision-free path is essential for any robot operating in a complex environment. In RoboCup 2024 adult-sized, autonomous humanoid soccer competition, 2 human referees, 4 robot handlers and 4 robots are constantly moving on a 14 m by 9 m field of which at times feels much smaller since everyone crowds the ball. This requires our robot, ARTEMIS, not only to plan obstacle-free paths in real time but also to re-plan and track effectively in spite of the vast amounts of noise, e.g., humans jumping in and out of frame. Our path planning and tracking process is shown in \cref{fig:complete_process}. Given an initial and final pose, the path planner finds the optimal path while avoiding obstacles using a modified version of visibility graphs called dynamic augmented visibility graphs (DAVG). Linear interpolation is then used to fill in the points with a time stamp. Finally, a modified version of Model Predictive Control (MPC), called Collision-Free MPC (cf-MPC), generates a control signal that ensures smooth, collision-free motion while considering the robot's dynamics. This planner played a big role in helping us score over 45 goals in 6 seated games in RoboCup 2024 and ultimately secure the championship title \cite{robocupRomelaSymposium}.



There are numerous methods for path planning: learning based methods \cite{raza2013path,reinhart2020learning,kulathunga2022reinforcement,popovic2024learning}, nonlinear optimization methods \cite{britzelmeier2023dynamic,zhang2023optimal,pardo2016evaluating}, control lyapunov function (CLF) and control barrier function (CBF) methods \cite{ames2016control,garg2019control,agrawal2017discrete,manjunath2021safe,tayal2024control,tayal2024polygonal}, heuristic methods like potential fields \cite{hwang1992potential,barraquand1992numerical,wang2000new,bounini2017modified,wu2023robot}, reachability methods \cite{liu2023radius,malone2014stochastic}, convex optimization with B\'ezier curves in Graphs of Convex Sets (GCS) \cite{marcucci2023motion,marcucci2024shortest}, rapidly-exploring random trees (RRT) \cite{lavalle2001randomized,wang2022path}, search based algorithms like A*\cite{jaiswal2022low}.


Path planning for a 35+ kg humanoid robot capable of running at 2 m/s must efficiently re-plan in a dynamic and complex environment, particularly when penalties apply for collisions with opponents. Moreover, computational efficiency is critical due to the hardware constraints and limited processing resources available on a humanoid robot.
To address these challenges, we chose a graph-based method, as it is computationally efficient and avoids local minima, ensuring a reliable long-term trajectory from start to finish.



Graphs have many different representations: uniform grids, log-polar grid which is non-uniform and denser around robots and obstacles \cite{jaiswal2022low,steffens2010multiresolution}, vertices of recursively divided blocks by quadtree \cite{lee2021visibility}, visibility graphs where the nodes are the vertices of the obstacles \cite{lozano1979algorithm,rohnert1986shortest,liu1995finding,huang2004dynamic} or waypoints of subotimal paths \cite{lee2021visibility}. Our proposed approach uses a type of visibility graph that chooses the vertices of the obstacles to be nodes. The shortest path in Euclidean space would be a straight line or along an edge of a convex obstacle. In our case we augment the states to account for turning since that slows our robot down. Our proposed method, DAVG, is further enhanced by using active regions and obstacles such that the entire graph does not need to be searched.



DAVG generates discrete points which we then interpolate between by assigning a time stamp to each point in the path. Since turning takes a longer than building momentum and running straight, we use linear interpolation, which is \(C_0\) continuous. This also maintains the shortest path property in Euclidean space. A potential added benefit is that when re-planning does occur the swings in changes between updated paths are reduced because of the lack of curvature.



MPC is popular for trajectory tracking \cite{kamel2017linear}. It can deal with nonlinear models and complex constraints directly without fine detailed modelling. In fact, for wheeled mobile robot models, which is nonholonomic, it is impossible to stabilize around the equilibrium point using smooth time-invarient feedback control \cite{brockett1983asymptotic,maurovic2011explicit,zhang2020trajectory}. In \cite{maurovic2011explicit}, to speed up the MPC, the authors described an explicit MPC where the optimal solution is calculated off-line and then used in the form of a look-up table. In \cite{zhang2020trajectory}, a nonlinear MPC (NMPC) with nonlinear obstacle avoidance constraints is introduced. 
It switches between tracking mode and obstacle avoidance mode using different objective functions. Our proposed method, cf-MPC, does not use switching for collision avoidance. cf-MPC has one problem formulation which allows for smoother transitions when balancing obstacle avoidance and robot dynamics. This played a very crucial role druing RoboCup when obstacles flashed in and out of our path and constraints. We can also linearize and simplify the constraints, and the formulation becomes a linear MPC (LMPC) which is a quadratic program (QP) problem, i.e., an efficiently solved problem. 



The following is a summary of contributions in this paper:
\begin{enumerate}
    \item Introduce DAVG which is capable of finding the shortest path efficiently while accounting for turning angles,
    \item Introduce cf-MPC which balances obstacles avoidance and robot dynamics into a single formulation in QP,
    \item Demonstrate its capabilities on hardware and in competition where we won RoboCup 2024.
\end{enumerate}

The article is organized as follows: in \cref{sec:path_planning} DAVG is detailed, 
in \cref{sec:traj_tracking} cf-MPC is covered, and in \cref{sec:experiment results} corresponding experiments and results are shown.

\section{Path Planning by DAVG}
\label{sec:path_planning}
\begin{figure}[t!]
    \centering
    \includegraphics[width=\linewidth]{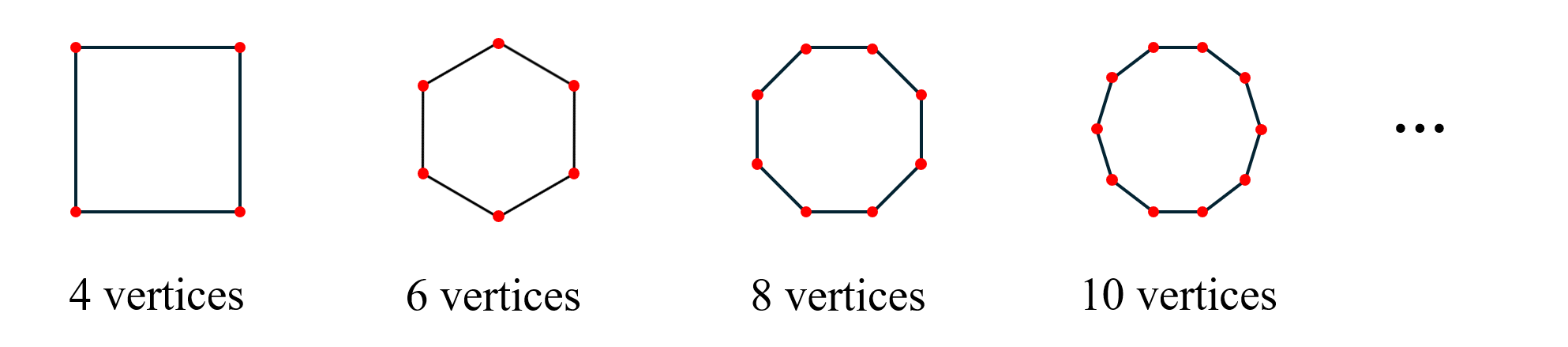}
    \caption{Regular polygon representation of obstacles.}
    \label{fig:regular_polygon_representation}
\end{figure}

Visibility graphs proposed by Lozano-P{\'e}rez and Wesley \cite{lozano1979algorithm} is a classical method of finding the shortest path of a point among polygonal obstacles. In our case, each obstacle (robot or person) is represented by a regular polygon \ref{fig:regular_polygon_representation}. 
\begin{figure}[t!]
    \centering
    \includegraphics[width=0.95\linewidth]{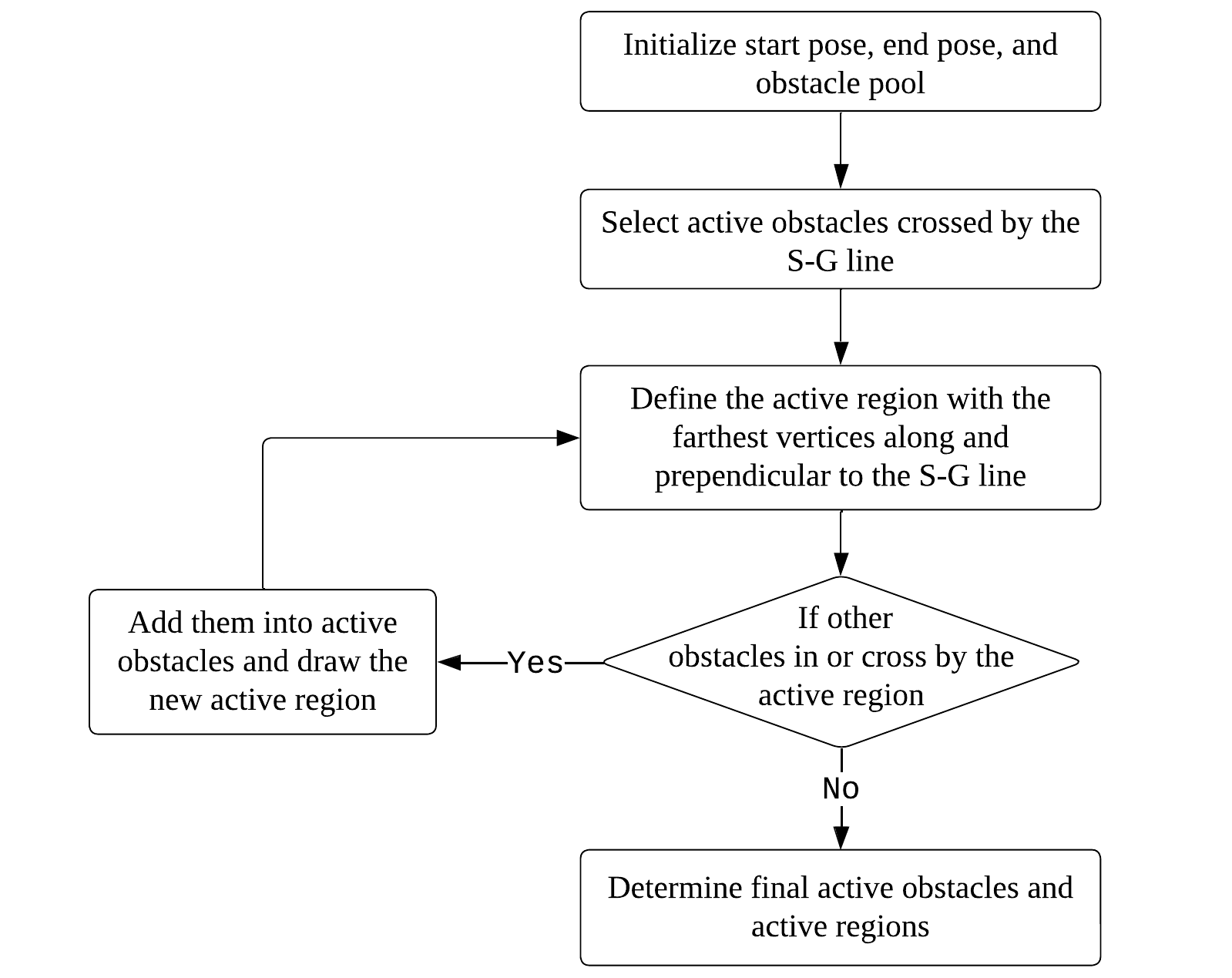}
    \caption{Process for selecting active obstacles and active regions.}
    \label{fig:active_region_process_chart}
\end{figure}
There are multiple methods to solve a visibility graph like A* \cite{warren1993fast}, Dijkstra's algorithm \cite{noto2000method,kang2008path}, or linear programming\cite{philpott1994continuous}.

\subsection{Preprocessing}
\begin{figure}[t!]
    \centering
    \includegraphics[width=0.95\linewidth]{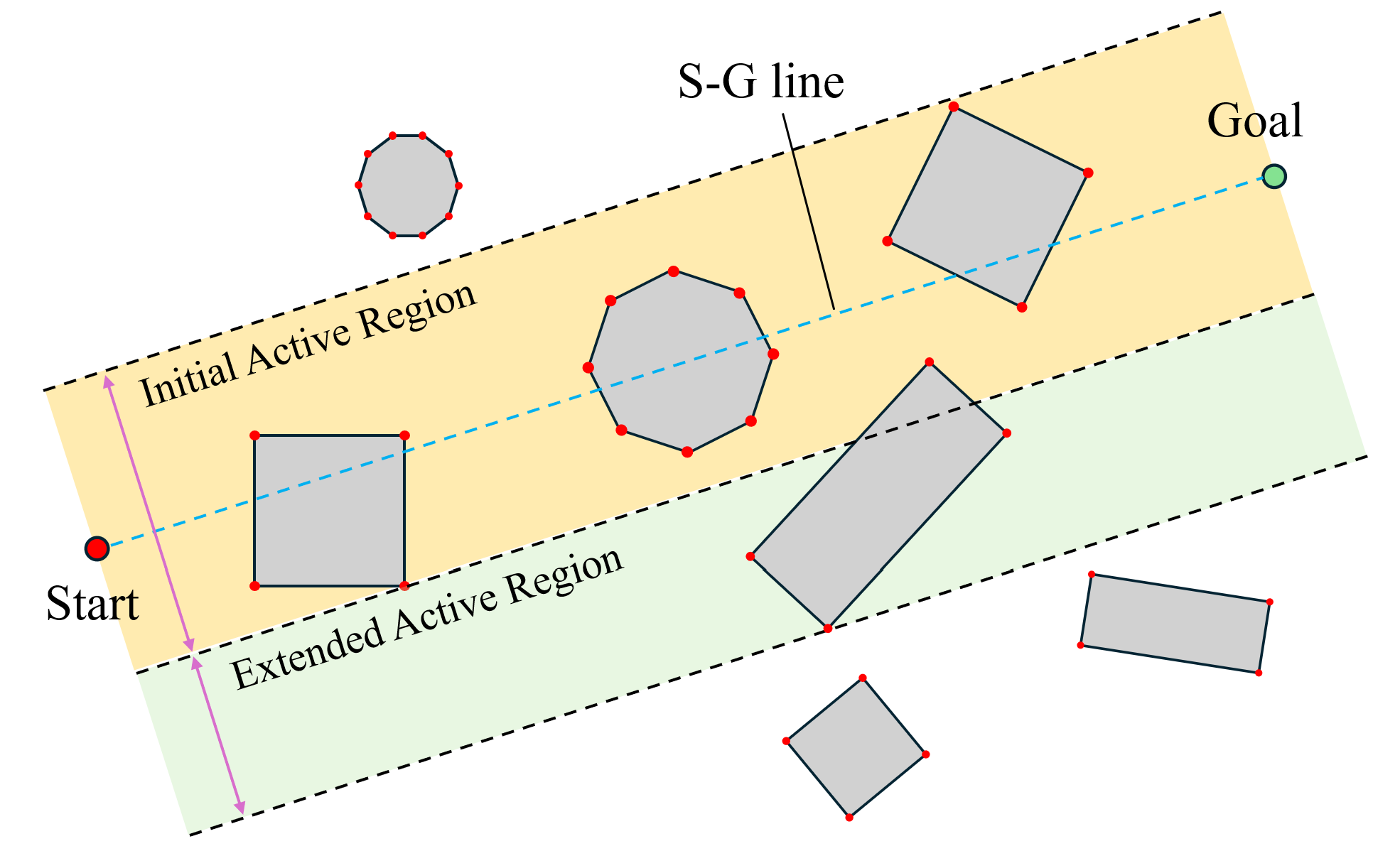}
    \caption{Illustration of selecting active region procedure.}
    \label{fig:active_region_demo}
\end{figure}

For complex environments, there may be many obstacles. Taking all vertices into account would be computational burdensome, and in fact some obstacles will not influence the optimal path.
To make the computational load more efficient, we only select necessary obstacles. Dynamic visibility graphs (DVG) 
by Huang and Chung \cite{huang2004dynamic} introduced a method to select active regions which contain only the necessary vertices and remove unnecessary ones. However, their method is complex when comparing the longest inner path and shortest outer path. In DAVG, we simplify this process. The preprocess procedures are shown in \cref{fig:active_region_process_chart}, and an example is given in \cref{fig:active_region_demo}. We define the straight line connecting the start and ending points as S-G line. A straight line is optimal in Euclidean space. A choice of a different space would result in a different optimal path absence of obstacles. Firstly, our algorithm finds the obstacles that intersect the S-G line, which forms the first active obstacles. These obstacles might influence the optimal trajectory. Then we find the farthest vertices along the S-G line and perpendicular to the S-G line. These define the length and width of the active region, which covers all the active obstacles. Then we check if there are other obstacles intersecting the active region. If there are, add these obstacles to the active obstacle since they might influence the optimal path and calculate the new active region. Repeat this until no obstacles intersect the active region, meaning that the obstacles outside of active region will not influence the optimal path. 

\begin{figure}[t!]
    \centering
    \includegraphics[width=0.95\linewidth]{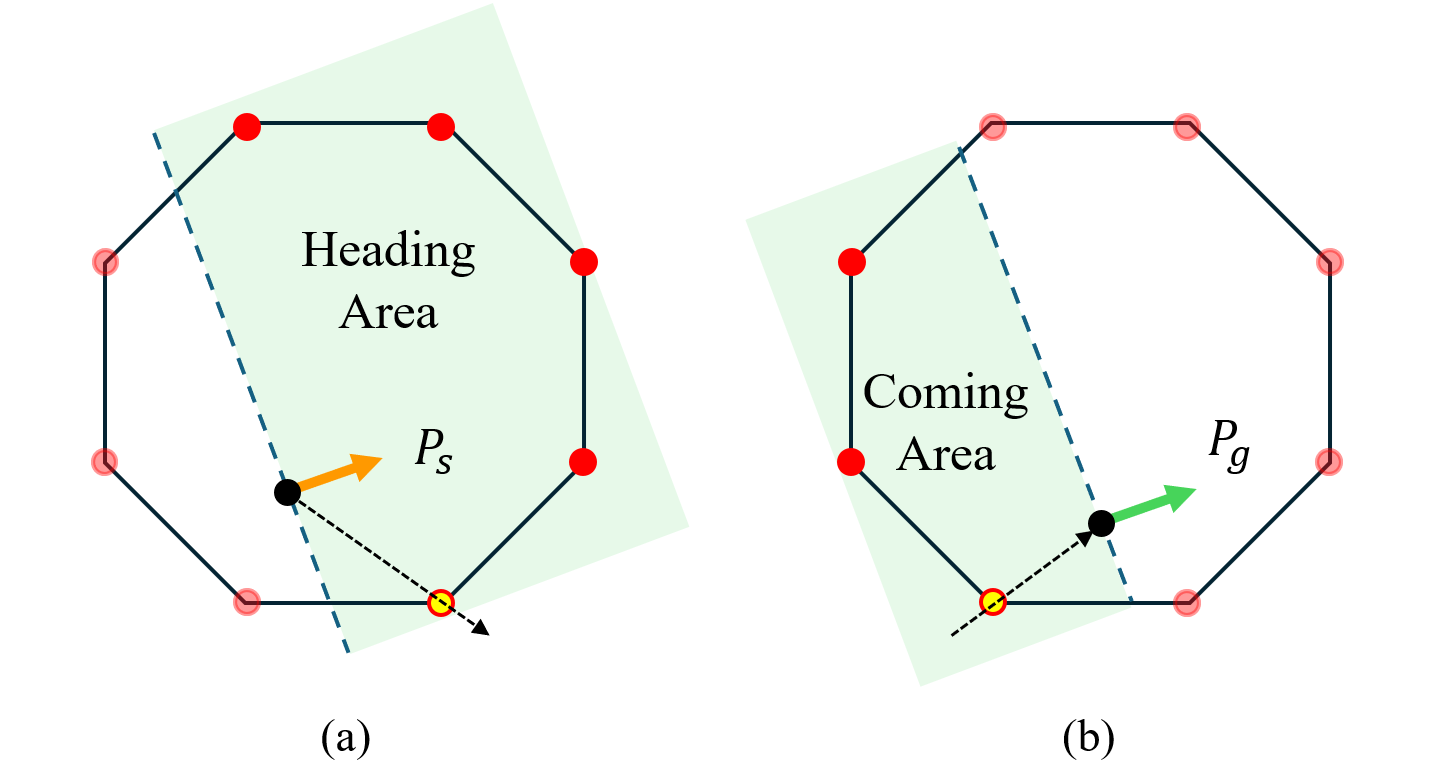}
    \caption{How to define the edge when the starting point or ending point is covered by an obstacle.}
    \label{fig:start_goal_point_edge}
\end{figure}

\begin{figure}[t!]
    \centering
    \includegraphics[width=0.93\linewidth]{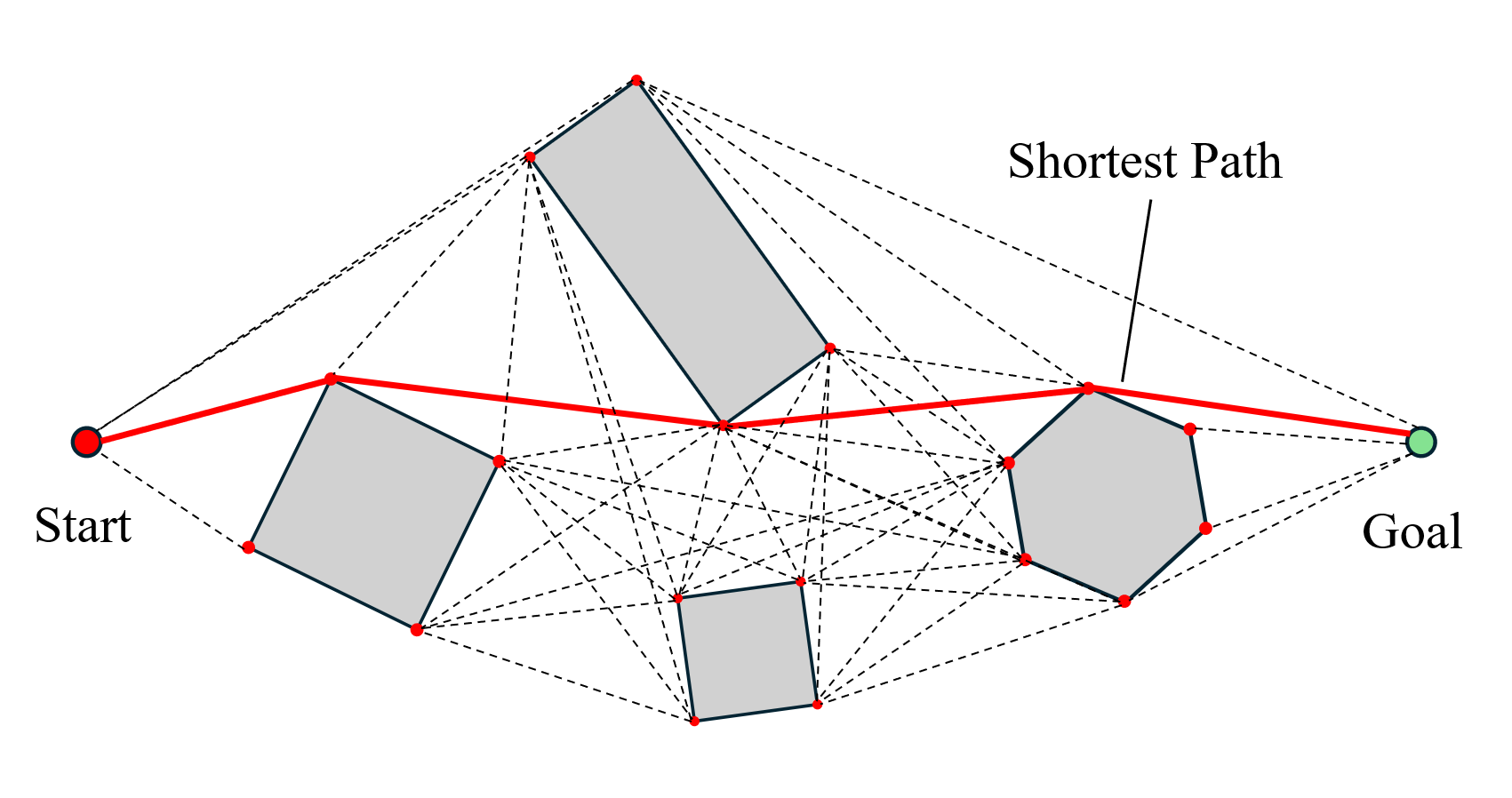}
    \caption{Finding the shortest path by our visibility graph called DAVG.}
    \label{fig:shortest_path_by_visibility_graph}
\end{figure}

\subsection{Shortest Path by Visibility Graph}

First we construct a visibility graph to find the shortest path that minimizes Euclidean distance. The nodes include the starting point $S$, goal point $G$ and vertices of the active obstacles $V_i$ which are not covered by other obstacles. There is an edge between two nodes if the edge does not have any intersections with obstacles. The weight of the edge connecting two nodes is the Euclidean distance. All such edges are generated.

If the starting point or the ending point is covered by an obstacle, they will not have an edge connecting them, and they are also not added to the node set $V$. We need to define vertices to go out of the obstacle in this case so that we can keep moving along similar directions. For starting points, the vertex should be in the \textit{heading} direction. For an obstructed goal point, the vertices should be in the \textit{coming} direction. If we are in the obstacles constraint, we want the path to go out of the obstacle as fast as possible as depicted in \cref{fig:start_goal_point_edge}. Therefore, the closest one is chosen. Finally, an edge exists between the chosen vertex and the starting point or the ending point. \cref{fig:shortest_path_by_visibility_graph} illustrates an example of finding the shortest path on such a graph.


\begin{algorithm}[t!]
\caption{Algorithm to Construct a Visibility Graph}
\label{Alg:Algorithm to Construct a Visibility Graph}
\begin{algorithmic}[1]

\State \textbf{Initialize the Graph:}
\State Let $VG = (V, E)$, where $V$ is the set of all nodes (start point $S$, goal point $G$, and obstacle vertices $V_i$), and $E$ is initially an empty set of edges.

\State \textbf{Remove invisible nodes:}
\ForAll{nodes in $V$}
    \If{node is covered by some obstacle}
        \State Remove it from the set $V$.
    \EndIf
\EndFor

\State \textbf{Define Edges by Looping Through All Pairs of Nodes:}
\ForAll{pairs of nodes $(V_i, V_j) \in V$}
    \State \textbf{Check Visibility:}
    \State Draw a line segment between $V_i$ and $V_j$.
    \If{line segment does not intersect any obstacle}
        \State Add edge $(V_i, V_j)$ to $E$ with distance weight. 
    \EndIf
\EndFor

\State \textbf{Define the Edge if $P_s$ or $P_g$ in Obstacle:}
\If{$P_s$ or $P_g$ is in obstacle}
    \State \textbf{Choose the leaving vertex on obstacle:}
    \If{$P_s$}
        \State The vertex should be in the going direction.
    \Else
        \State The vertex should be in the entering direction.
    \EndIf
    \State Choose the closest vertex.
    \State Define edge between chosen vertex and $P_s$ or $P_g$.
\EndIf

\end{algorithmic}
\end{algorithm}

\subsection{Shortest Path with Turning Angle}
\begin{figure}[t!]
    \centering
    \includegraphics[width=\linewidth]{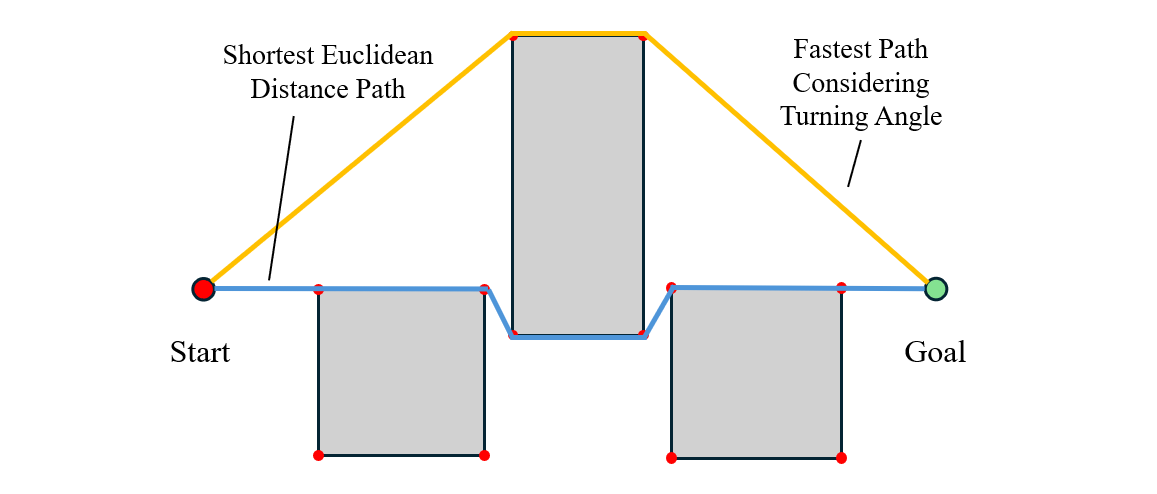}
    \caption{How the augmented visibility graph gives a different path by augmenting the state with a weight determined by the turning angle.}
    \label{fig:augmented_visibility_graph_demo}
\end{figure}

A visibility graph does not inherently account for constraints like minimizing the turning angles along the path. The shortest path found by a visibility graph might have sharp turns, especially, around obstacle corners, since it connects visible vertices with straight lines. For humanoid robots and many other systems turning is time-consuming. Therefore, it is preferable to take the turning angle into consideration when constructing these graphs. One way to achieve this is to augment the states.
The current weight of the edge between two states depends on both the Euclidean distance and turning angle. We call this modified graph an augmented visibility graph (AVG). \cref{fig:augmented_visibility_graph_demo} depicts an example by which the path is altered by using AVG for turning angles. Turning angles are added in as additional weights on the edges of the graph.
For humanoid robots, turning can be slow, and, therefore, the outer path is chosen, though longer, reduces the amount of turning. The combination of active regions and augmented edges defines DAVG.

\begin{algorithm}[t!]
\caption{Augmented Visibility Graph}
\label{Alg:Algorithm to Construct an Augmented Visibility Graph}
\begin{algorithmic}[1]

\State \textbf{Initialize the Graph:}
\State Let $AVG = (\hat{V}, \hat{E})$, where $\hat{V}$ is the set of all combined state nodes, and $\hat{E}$ is initially an empty set of edges.

\State \textbf{Construct the normal Visibility Graph} $VG = (V, E)$ as described in Algorithm~\ref{Alg:Algorithm to Construct a Visibility Graph}.

\State \textbf{Define the Combined State:}
\ForAll{edges $(V_i, V_j) \in E$}
    \State Represent combined state as $\hat{V_{ij}} = (V_i, V_j)$, where:
    \State \hspace{1em} $V_i$ represents the past vertex.
    \State \hspace{1em} $V_j$ represents the current vertex.
    \State Add $\hat{V_{ij}}$ to the set $\hat{V}$.
\EndFor
\State Add starting state $\hat{SS}$ and goal state $\hat{GG}$ to $\hat{V}$.

\State \textbf{Define Edges and Weights with Combined States:}
\ForAll{nodes $\hat{V_{ij}}$ and $\hat{V_{jk}} \in \hat{V}$}
    \If{an edge exists between $V_i$ and $V_j$}
        \State Add edge $\hat{E_{ijk}}$ to $\hat{E}$, connecting $\hat{V_{ij}}$ and $\hat{V_{jk}}$.
        \State Weight of $\hat{E_{ijk}}$ combines Euclidean distance $E_{jk}$ 
        \State and turning angle from $E_{ij}$ to $E_{jk}$.
    \EndIf
\EndFor

\end{algorithmic}
\end{algorithm}

\section{trajectory tracking}
\label{sec:traj_tracking}
Interpolating the results from DAVG, we get a trajectory which avoids obstacles. This section shows how to track that trajectory with an MPC. Note that even though we can guarantee that the trajectory from DAVG will avoid obstacles in theory, the robot can still collide with obstacles if it cannot track its trajectory well. In some cases the robot cannot properly track its trajectory due to its dynamic and physical limitations, e.g., speed and acceleration limits. Therefore, we introduce a collision free constraint to the MPC to help the robot avoid obstacles. The constraint can be relaxed to ensure a feasible solutions while trying to avoid the obstacle. This provides a method for short-term planning. 

\subsection{Dynamic Model}
The 2D pose on the field is $\mathbf{p} :=
\begin{bmatrix}
x &
y &
\theta
\end{bmatrix}^T$, where \(x\in \mathbb{R},y\in \mathbb{R} \) is the position and $\theta \in (-\pi, \pi]$ is the orientation, shown in \cref{fig:2D_model_vx_vy_w}.
For a humanoid, it is capable of moving both forward and sideways. In the analysis below, we are using the humanoid model. However, in fact, the ability for a humanoid robot to move sideways might be poor. Therefore, it is acceptable to use a model for mobile robots for simplicity. The discretized model we used is as follows:

\begin{figure}[t!]
    \centering
    \includegraphics[width=0.9\linewidth]{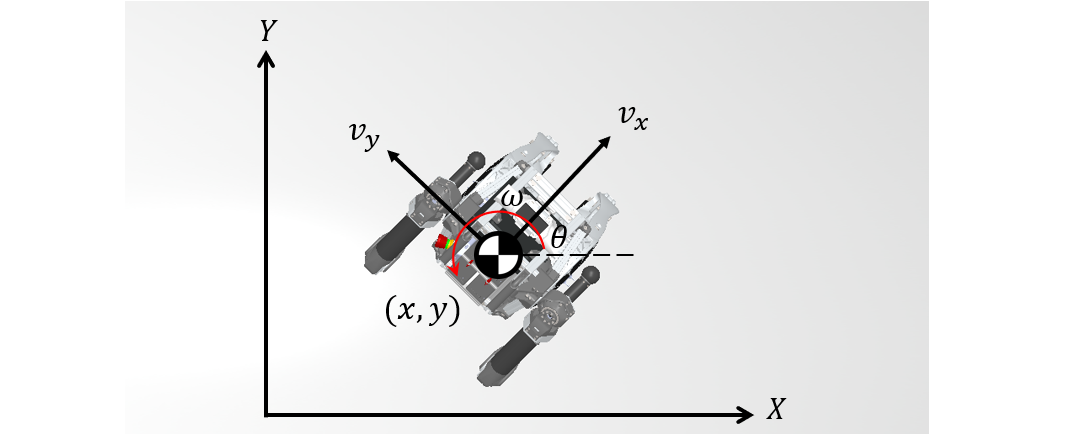}
    \caption{Coordinate system for our humanoid robot.}
    \label{fig:2D_model_vx_vy_w}
\end{figure}
\begin{equation*}
\begin{bmatrix}
x_{k+1} \\
y_{k+1} \\
\theta_{k+1}
\end{bmatrix}
=
\begin{bmatrix}
x_{k} \\
y_{k} \\
\theta_{k}
\end{bmatrix}
+
\begin{bmatrix}
\cos(\theta_{k}) & -\sin(\theta_{k}) & 0 \\
\sin(\theta_{k}) & \cos(\theta_{k}) & 0 \\
 0 & 0 & 1
\end{bmatrix}
\begin{bmatrix}
v_x \\
v_y \\
\omega
\end{bmatrix}
dt
\end{equation*}
where $dt$ is the time step. Define the state \(X\) and control input \(u\) as : $\mathbf{X_k} :=
\begin{bmatrix}
x_k & y_k & \theta_k
\end{bmatrix}^T$ 
and 
$\mathbf{u_k} :=
\begin{bmatrix}
v_{xk} & v_{yk} & \omega_{k}
\end{bmatrix}^T$.
The system is as follows: 
\begin{equation*}
X_{k+1} = f_d(X_k, u_k)
\end{equation*}

\subsection{Nonlinear Model Predictive Control}
In NMPC, we predict the next $N$ steps, and the objective function is formulated to track its reference trajectory. Considering the discrete dynamics of a humanoid robot, the objective function for the NMPC is as follows:
\begin{align*}
J_0 = \sum_{i=1}^{N} (X_{i} - X_{r,i})^T Q (X_{i} - X_{r,i}) + \sum_{i=0}^{N-1}  u_{i}^T R u_{i} 
\label{eq:J_cost1}
\end{align*}
where \(X_{r,i}\) is the reference state at time step \(i\). \(Q \in \mathbb{R}^{n \times n}\) is a weight matrix, and \(R \in \mathbb{R}^{m \times m}\) is a weight matrix. 
The reference trajectory follows:
\begin{equation*}
X_{r,k+1} = f_d(X_{r,k}, u_{r,k})
\end{equation*}
The initial point is $X_0$, and its dynamic constraint is as follows:
\begin{equation*}
x_{i+1} = f_d(x_{i}, u_{i}), \quad i = 0, \ldots, N-1
\end{equation*}
As we define a circular region that we do not want the trajectory to violate, the collision free constraint is as follows:
\begin{align*}
(x_i - x_{\text{obs}, j})^2 + (y_i - y_{\text{obs}, j})^2 &\geq R^2, 
&\quad \forall \substack{i = 1, \ldots, N \\ j = 1, \ldots, k}
\end{align*}
We can add a positive slack variable \(\delta_j \in \mathbb{R}^+
\) to relax the hard collision free constraint to make sure that the problem is feasible. 
\begin{align*}
(x_i - x_{\text{obs}, j})^2 + (y_i - y_{\text{obs}, j})^2 + \delta_j &\geq R^2, 
&\quad \forall \substack{i = 1, \ldots, N \\ j = 1, \ldots, k} \\
\delta_j &\geq 0
\end{align*}
The objective function becomes:
\begin{align}
J = J_0 + \rho \sum_{j = 1}^{k} \delta_j^2 
\label{eq:J}
\end{align}
where \(\rho\) is the weight on the collision free constraint. Larger \(\rho\) makes the constraint stricter. We can also add a speed constraint:
\begin{align}
\begin{bmatrix}
    v_{\text{xmin}} \\
    v_{\text{ymin}} \\
    \omega_{\text{min}}
\end{bmatrix}
\leq
\begin{bmatrix}
    v_{x_k} \\
    v_{y_k} \\
    \omega_{k}
\end{bmatrix}
\leq
\begin{bmatrix}
    v_{\text{xmax}} \\
    v_{\text{ymax}} \\
    \omega_{\text{max}}
\end{bmatrix}
\label{eq:speed_constraint}
\end{align}
This is a summary of the NMPC formulation:

\begin{align*}
\min_{\substack{ X_{1:N} \\ u_{0:N-1}}} & \; 
J \text{ from } \cref{eq:J}\\
\text{s.t.} \quad & X_{r,k+1} = f_d(X_{r,k}, u_{r,k}) \\
& (x_k - x_{\text{obs}, j})^2 + (y_k - y_{\text{obs}, j})^2 + \delta_j \geq R^2 \\
& \delta_j \geq 0 \\
& 
\text{Speed constraint from \cref{eq:speed_constraint}}
\end{align*}


\subsection{Linear Model Predictive Control}
To make the optimization faster, the dynamic model can be linearized around the trajectory. 
We compute the Jacobian matrices of the nonlinear discrete-time dynamics with respect to the state and input at the current linearization point:
\[
A_k = \left. \frac{\partial f_d(X, u)}{\partial X} \right|_{\substack{X = X_{r,k} \\ u = u_{r,k}}} \qquad
B_k = \left. \frac{\partial f_d(X, u)}{\partial u} \right|_{\substack{X = X_{r,k} \\ u = u_{r,k}}}
\]
where: \(A_k \in \mathbb{R}^{n \times n}\) is the Jacobian matrix with respect to the state at time step \(k\), and \(B_k \in \mathbb{R}^{n \times m}\) is the Jacobian matrix with respect to the input at time step \(k\).
Therefore, the dynamic constraints is as follows:
\begin{equation*}
    X_{k+1} = X_{k} + A_k  (X_{k}-X_{r,k}) + B_k  (u_k - u_{r,k})
\end{equation*}

\begin{figure}[t!]
    \centering
    \includegraphics[width=\linewidth]{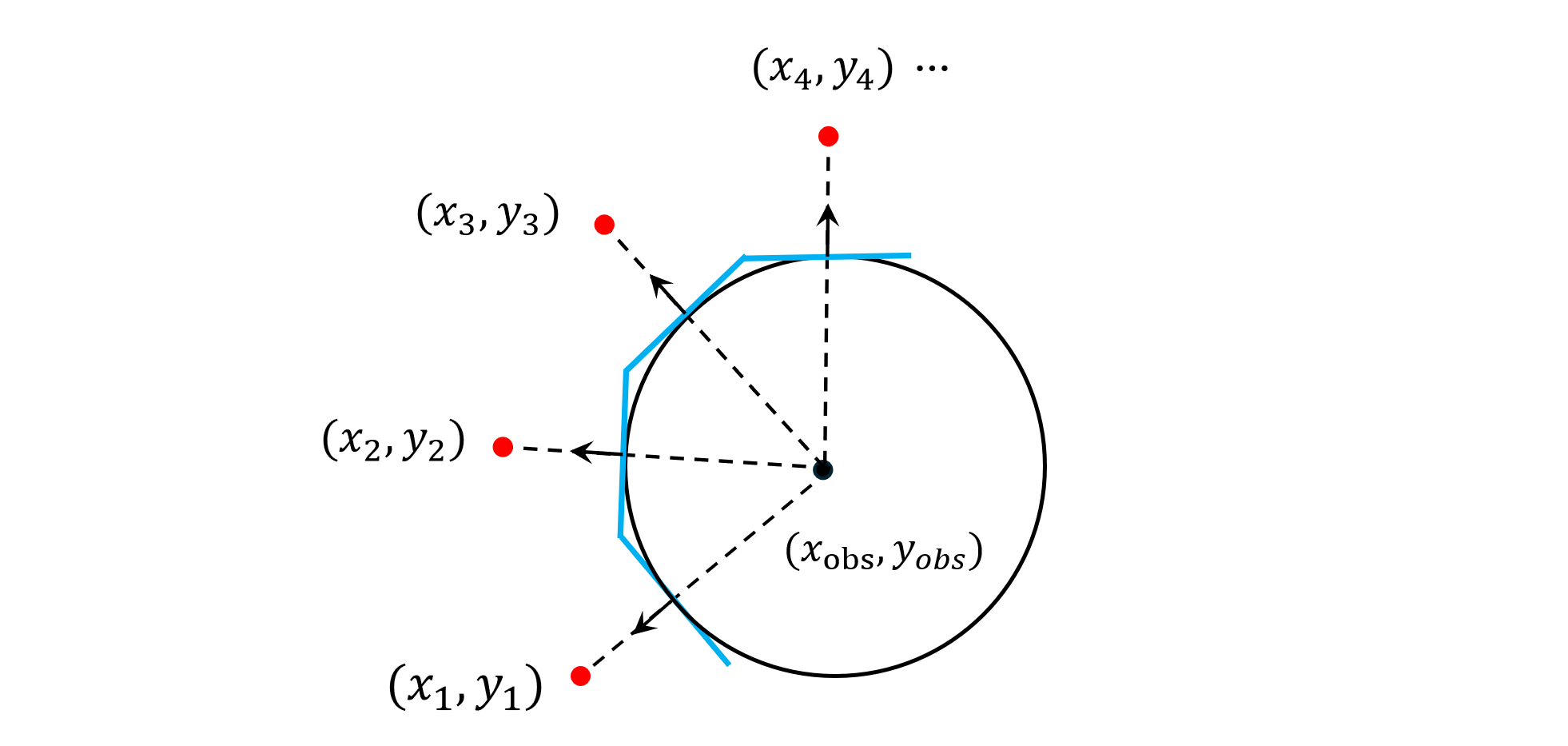}
    \caption{Linearization of collision free constraints.}
    \label{fig:linearization_of_collision_free_constraint}
\end{figure}

Additionally, the collision free constraint can also be simplified by linear constraints. The obstacle circle can be replaced by its tangent plane as shown in \cref{fig:linearization_of_collision_free_constraint}.
Define \(V_{k,j} = \begin{bmatrix}
        x_{r,k} - x_{\text{obs},j} \\
        y_{r,k} - y_{\text{obs},j}
    \end{bmatrix} \), 
then the collision free constraint becomes:
\begin{align*}
    &\begin{bmatrix}
        x_{k} - x_{\text{obs},j} & y_{k} - y_{\text{obs},j}
    \end{bmatrix}
    V_{k,j} \notag 
    &\geq \left\lVert
    V_{k,j}
    \right\rVert 
    (R_{\text{obs},j} - \delta_j)
\end{align*}
\begin{figure}[t!]
    \centering
    \includegraphics[width=0.9\linewidth]{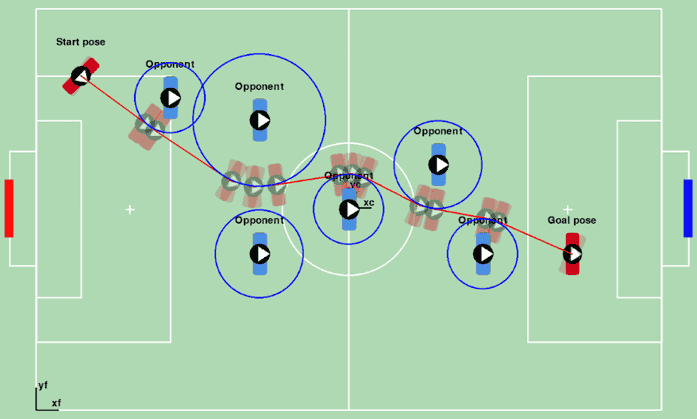}
    \caption{Diagram of the optimal path of DAVG.}
    \label{fig:lp_sim_figure}
\end{figure}
\begin{figure}[t!]
    \centering
    \includegraphics[width=0.9\linewidth]{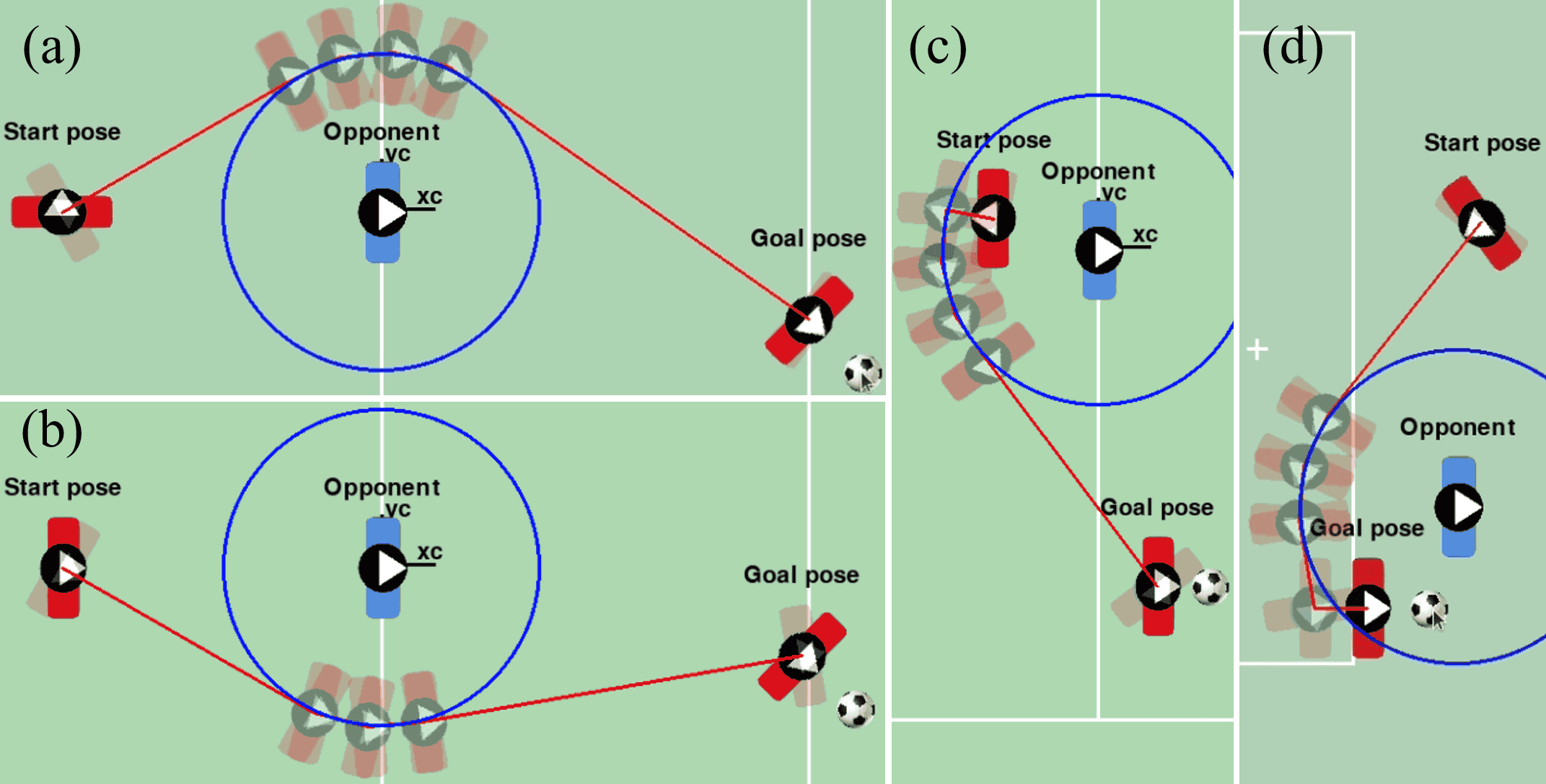}
    \caption{Diagram of specific cases of DAVG: (a) and (b) show how the turning angle affects the choice of the optimal path in simulation; (c) is an example of finding a vertex to exit an obstacle from an internal starting point; and (d) is an example of finding a vertex to exit an obstacle from an internal ending point.}
    \label{fig:lp_edge_case_figure}
\end{figure}
\begin{figure}[t!]
    \centering
    \includegraphics[width=0.9\linewidth]{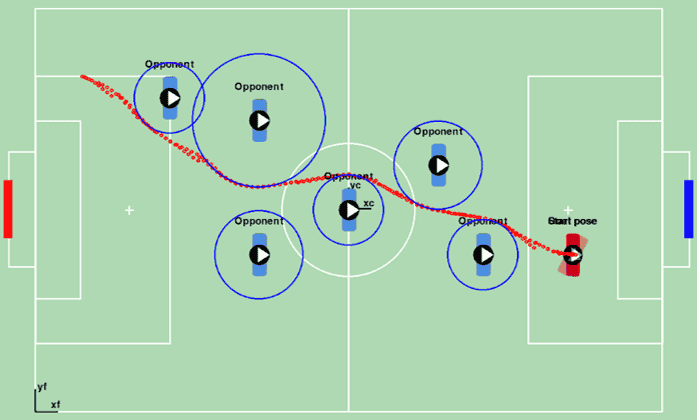}
    \caption{Diagram of our linear cf-MPC tracking the calculated path in \ref{fig:lp_sim_figure}.}
    \label{fig:mpc_tracking_sim}
\end{figure}
It is a linear constraint, where \(R_{obs,j}\) is the radius of obstacle \(j\), and \(\delta_j\) is a slack variable for obstacle \(j\) to relax the constraint and guarantee feasibility.
Therefore, in summary, the LMPC can be formulated as a QP problem in the following:
\begin{align*}
\min_{\substack{ X_{1:N} \\ u_{0:N-1}}} & \; 
J \text{ cost from } \cref{eq:J}
\\
\text{s.t.} \quad & X_{r,k+1} = f_d(X_{r,k}, u_{r,k})\\
&\begin{bmatrix}
        x_{k} - x_{\text{obs},j} & y_{k} - y_{\text{obs},j}
    \end{bmatrix}
    V_{k,j} 
    \geq \left\lVert
    V_{k,j}
    \right\rVert 
    (R_{\text{obs},j} - \delta_j) \\
& \delta_j \geq 0  \\
& 
\text{Speed constraint from \cref{eq:speed_constraint}}
\end{align*}

\section{Results} 
\label{sec:experiment results}
In this section, we present both the simulation and hardware implementation of our proposed path planning and trajectory tracking method. We also encourage readers to view the online competition videos to evaluate our performance. For the simulation, we modeled a soccer field in Python to validate our approach. The nonlinear problem was solved using SNOPT, and the linear problem with Gurobi. All code was executed on an HP Elite Mini 800 G9, equipped with an Intel Core i7-12700T processor and an Nvidia GeForce RTX 3050 Ti 4GB GPU.

\subsection{Path Planning Simulation}
First, we present a simulation result of the optimal path found by our DAVG method, with 6 obstacles of varying sizes present on the field, as shown in the \cref{fig:lp_sim_figure}. Each obstacle is represented by a polygon with 18 sides. The red poses with higher transparency around obstacles represents vertices on the path. 




We then tested some specific properties of the DAVG method. \cref{fig:lp_edge_case_figure} (a) and (b) shows how the augmented visibility graph generates different paths by giving different starting angles. \cref{fig:lp_edge_case_figure} (c) and (d) demonstrates our method successfully finding optimal vertices to exit when the starting or ending point is covered by an obstacle.

\subsection{MPC Simulation}
We validated the LMPC performance before implementing it on hardware. Ten-step horizons are predicted at each time step in our LMPC, with each time step fixed at 0.25s. The actual control output is shown in \cref{fig:mpc_tracking_sim}. It demonstrates good tracking performance. This result highlights the effectiveness of our framework and its readiness for implementation on the actual hardware platform for competition.




\subsection{Hardware Implementation}
\begin{figure}[t!]
    \centering
    \includegraphics[width=0.9\linewidth]{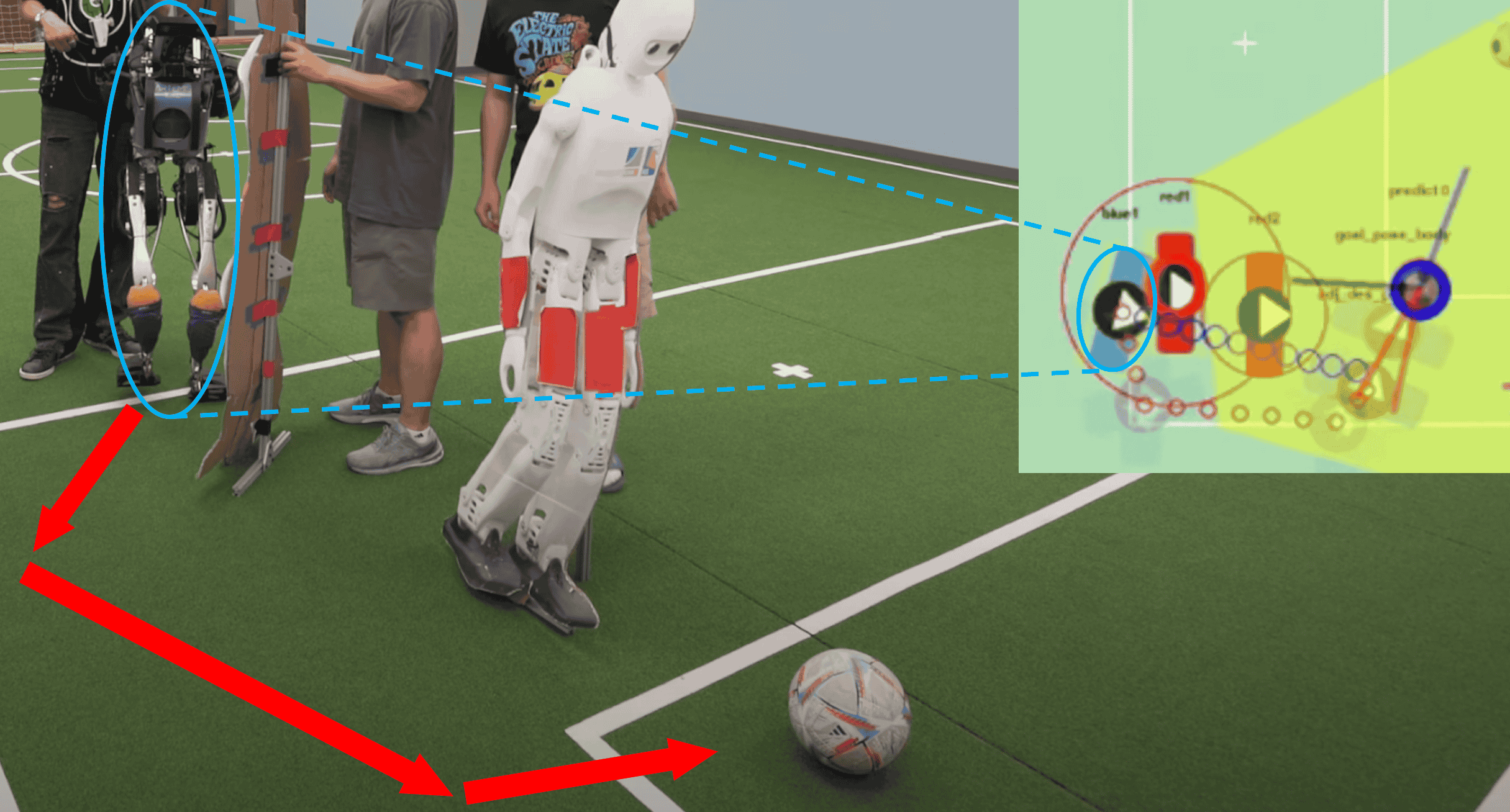}
    \caption{Field test on ARTEMIS with the path visualization when obstacles flash suddenly into the path. Red dots on the visualizer represent the new path while blue dots represent the old path.}
    \label{fig:field_test_fig}
\end{figure}

\begin{figure}[t!]
    \centering
    \includegraphics[width=0.97\linewidth]{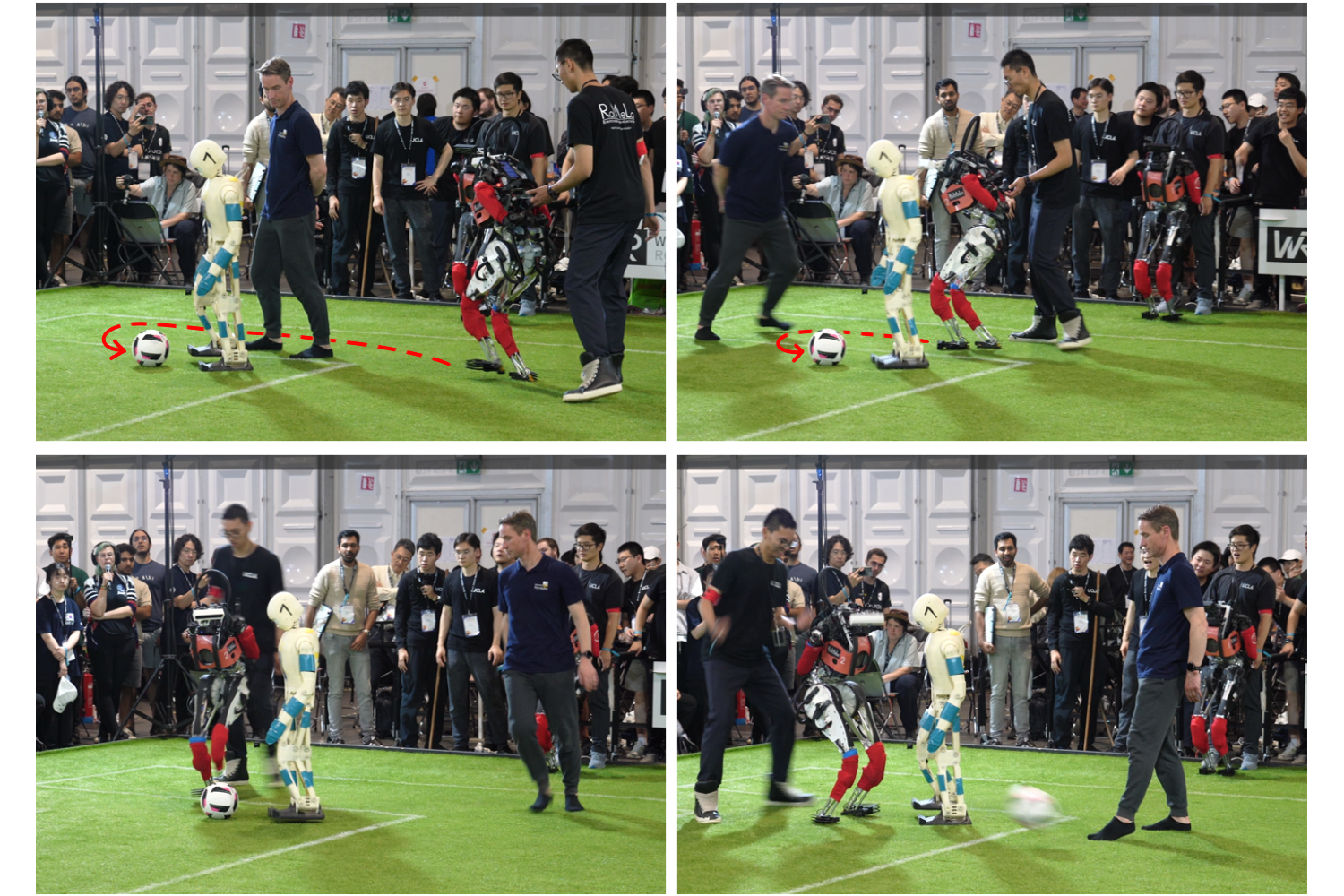}
    \caption{ARTEMIS successfully bypasses the opponent robot in the competition and retrieves the ball. 
    }
    \label{fig:competition_figure}
\end{figure}

We have also demonstrated the feasibility of our proposed control framework through hardware experiments using our humanoid robot, ARTEMIS. The robot's location is determined using the localization methods we have proposed \cite{hou2025localization}. 
\cref{fig:field_test_fig} shows a snapshot of the experiment where obstacles were suddenly placed in front of ARTEMIS. The adjusted path, even when ARTEMIS was already within the circular constraint region of the obstacle, illustrates the framework’s capability to avoid sudden obstacles. \cref{fig:competition_figure} shows snapshots of ARTEMIS successfully maneuvering around an opponent moving along a straight path to the soccer ball, retrieving it, and clearing it from the penalty area for defensive play.

\subsection{Solving Time}
The DAVG solving time is mainly influenced by the number of obstacles and the number of arcs for each polygon. \cref{tab:DAVG_Solving_Time} shows the average solving time of DAVG.

\begin{table}[htbp]
\centering
\caption{DAVG Solving Time}
\begin{tabular}{|c|c|c|c|c|c|c|c|c|}
\hline
\multicolumn{1}{|c|}{\textbf{N\_obs}} & \multicolumn{2}{c|}{2} & \multicolumn{2}{c|}{4} & \multicolumn{2}{c|}{6} & \multicolumn{2}{c|}{8} \\ \hline
\textbf{N\_arc} & 4 & 10 & 4 & 10 & 4 & 10 & 4 & 10 \\ \hline
\textbf{(ms)} & 0.33 & 1.1 & 0.53 & 2.1 & 0.77 & 4.3 & 1.41 & 8.0 \\ \hline
\end{tabular}
\label{tab:DAVG_Solving_Time}
\end{table}

The MPC solving time is mainly influenced by the prediction horizon.
\cref{tab:LMPC_and_NMPC_solving_time} shows the average solving time of LMPC and NMPC with two obstacles in the way.  
\begin{table}[htbp]
\centering
\caption{LMPC and NMPC average solving time}
\begin{tabular}{|c|c|c|c|c|}
\hline
\textbf{N\_step} & \textbf{5} & \textbf{10} & \textbf{15} & \textbf{20} \\
\hline
\textbf{LMPC (ms)} & 0.65 & 0.88 & 1.1 & 1.4 \\
\hline
\textbf{NMPC (ms)} & 4.2 & 15.8 & 40.6 & 84.6 \\
\hline
\end{tabular}
\label{tab:LMPC_and_NMPC_solving_time}
\end{table}



In the competition, the path planning and tracking process runs at 80–100 Hz using NMPC and around 400 Hz using LMPC, while also handling other computationally intensive tasks such as vision, localization, locomotion, and high-level decision-making on the same computer.

\section{Conclusion} 
\label{sec:conclusion}
In this paper, we proposed DAVG which is capable of finding the shortest path efficiently while taking turning angle into account by augmenting states, since for our humanoid robot turning is more difficult and time consuming. To track this path we proposed cf-MPC which balances avoiding obstacles and robot dynamics into a single formulation as opposed to other switching methods. 
This provides for a smooth transition between tracking and obstacle avoidance which is necessary in such a noisy environment. The combination of both DAVG and cf-MPC allow for a good combination of long-term and short-term planning while maintaining reasonable solve speeds, approximately 100 Hz for NMPC and 400 Hz for LMPC. Though our examples are in 2D, it can be easily extend to higher dimensions \cite{lozano1979algorithm}. Performance and speed 
were verified in both simulation and hardware during the competition, which ultimately helped us win RoboCup 2024 by a wide margin.


\section*{Acknowledgements}
We would like to thank everyone who participated in RoboCup 2024 AdultSize humanoid robot soccer competition, and our teammates and past contributors: Xuan Lin, Colin Togashi, Aditya Navghare, Alvin Zhu, Arturo Flores Alvarez, Edmond Wang, Ethan Hong, Quanyou Wang, Hyunwoo Nam, Kyle Gillespie, Yeting Liu, Alex Xu, Yicheng Wang, Luke Mariak, Min Sung Ahn, Taoyuanmin Zhu, Junjie Shen, Jingwen Zhang, Ji Sung Ahn, and Justin Quan,.




\end{document}